%% file: main.tex
\documentclass[10pt,twocolumn,letterpaper]{article}

\usepackage{cvpr}              %

\usepackage{graphicx}
\usepackage{amsmath}
\usepackage{amssymb}
\usepackage{booktabs}

\usepackage[accsupp]{axessibility}  %

\usepackage[pagebackref,breaklinks,colorlinks]{hyperref}

\usepackage[capitalize]{cleveref}
\crefname{section}{Sec.}{Secs.}
\Crefname{section}{Section}{Sections}
\Crefname{table}{Table}{Tables}
\crefname{table}{Tab.}{Tabs.}

\def\cvprPaperID{644} %
\def\confName{CVPR}
\def\confYear{2023}

\input{packages}

\input{macro}

\begin{document}

\title{3D-aware Conditional Image Synthesis}

\author{Kangle Deng \quad Gengshan Yang \quad Deva Ramanan \quad Jun-Yan Zhu \\
Carnegie Mellon University 
}
\vspace{-20pt}

\input{figTex/fig_teaser}

\input{sections/00_abstract}
\input{sections/01_introduction}

\input{sections/02_related_work}

\input{sections/03_method}

\input{sections/04_experiment}
\input{sections/05_discussion}

{\small
\bibliographystyle{ieee_fullname}
\bibliography{main}
}

\clearpage
\appendix
\input{sections/06_appendix.tex}

\end{document}

%% file: packages.tex
\usepackage{times}
\usepackage{epsfig}
\usepackage{graphicx}
\usepackage{amsmath}
\usepackage{amssymb}
\usepackage{color,xcolor}

\usepackage{comment}

\usepackage{pifont}

\usepackage{microtype}
\usepackage[font=small]{caption}
\usepackage{arydshln}
\usepackage{tabularx}
\usepackage{adjustbox}
\usepackage{array}
\usepackage{booktabs}
\usepackage{colortbl}
\usepackage{float,wrapfig}
\usepackage{hhline}
\usepackage{multirow}
\usepackage{subcaption} %
\usepackage[percent]{overpic}

\usepackage{stmaryrd}

\usepackage{amsmath,amsfonts,amssymb}
\usepackage{bm}
\usepackage{nicefrac}
\usepackage{microtype}
\usepackage{dsfont}
\usepackage{changepage}
\usepackage{extramarks}
\usepackage{fancyhdr}
\usepackage{setspace}
\usepackage{soul}
\usepackage{xspace}

\usepackage{algorithm, algorithmic}
\usepackage{enumitem}
\usepackage[title]{appendix}

\usepackage{duckuments}
\usepackage{subcaption}

%% file: macro.tex
\usepackage{fp}
\usepackage{expl3}[2012-07-08]
\ExplSyntaxOn
\cs_new_eq:NN \fpeval \fp_eval:n
\ExplSyntaxOff

\usepackage{amsmath,amsfonts,bm}

\def\x{{x}}

\def\xi{{\x_i}}

\newcommand{\reffig}[1]{Figure~\ref{fig:#1}}
\newcommand{\refsec}[1]{Section~\ref{sec:#1}}
\newcommand{\refapp}[1]{Appendix~\ref{sec:#1}}
\newcommand{\reftbl}[1]{Table~\ref{tbl:#1}}

\newcommand{\refeq}[1]{Equation~\ref{eq:#1}}

\newcommand{\lblfig}[1]{\label{fig:#1}}
\newcommand{\lblsec}[1]{\label{sec:#1}}
\newcommand{\lbleq}[1]{\label{eq:#1}}
\newcommand{\lbltbl}[1]{\label{tbl:#1}}

\newcommand{\ignorethis}[1]{}
\newcommand{\revision}[1]{\color{black}#1\color{black}}
\newcommand{\myparagraph}[1]{\smallskip \noindent \textbf{#1}}

\def\eqref#1{equation~\ref{#1}}

\def\1{\bm{1}}

\DeclareMathAlphabet{\mathsfit}{\encodingdefault}{\sfdefault}{m}{sl}
\SetMathAlphabet{\mathsfit}{bold}{\encodingdefault}{\sfdefault}{bx}{n}

\newcommand{\ignore}[1]{}

\makeatletter
\DeclareRobustCommand\onedot{\futurelet\@let@token\@onedot}
\def\@onedot{\ifx\@let@token.\else.\null\fi\xspace}

\makeatother

\definecolor{MyDarkBlue}{rgb}{0,0.08,1}
\definecolor{MyDarkGreen}{rgb}{0.02,0.6,0.02}
\definecolor{MyDarkRed}{rgb}{0.8,0.02,0.02}
\definecolor{MyDarkOrange}{rgb}{0.40,0.2,0.02}
\definecolor{MyPurple}{RGB}{111,0,255}
\definecolor{MyRed}{rgb}{1.0,0.0,0.0}
\definecolor{MyGold}{rgb}{0.75,0.6,0.12}
\definecolor{MyDarkgray}{rgb}{0.66, 0.66, 0.66}
\definecolor{myorange}{RGB}{255,69,0}

\definecolor{revisionColor}{RGB}{255,69,0}

\newcommand{\ourmethod}{\texttt{pix2pix3D}\xspace}

%% file: figTex/fig_teaser.tex
\twocolumn[{%
\maketitle
\centering
\includegraphics[width=\linewidth]{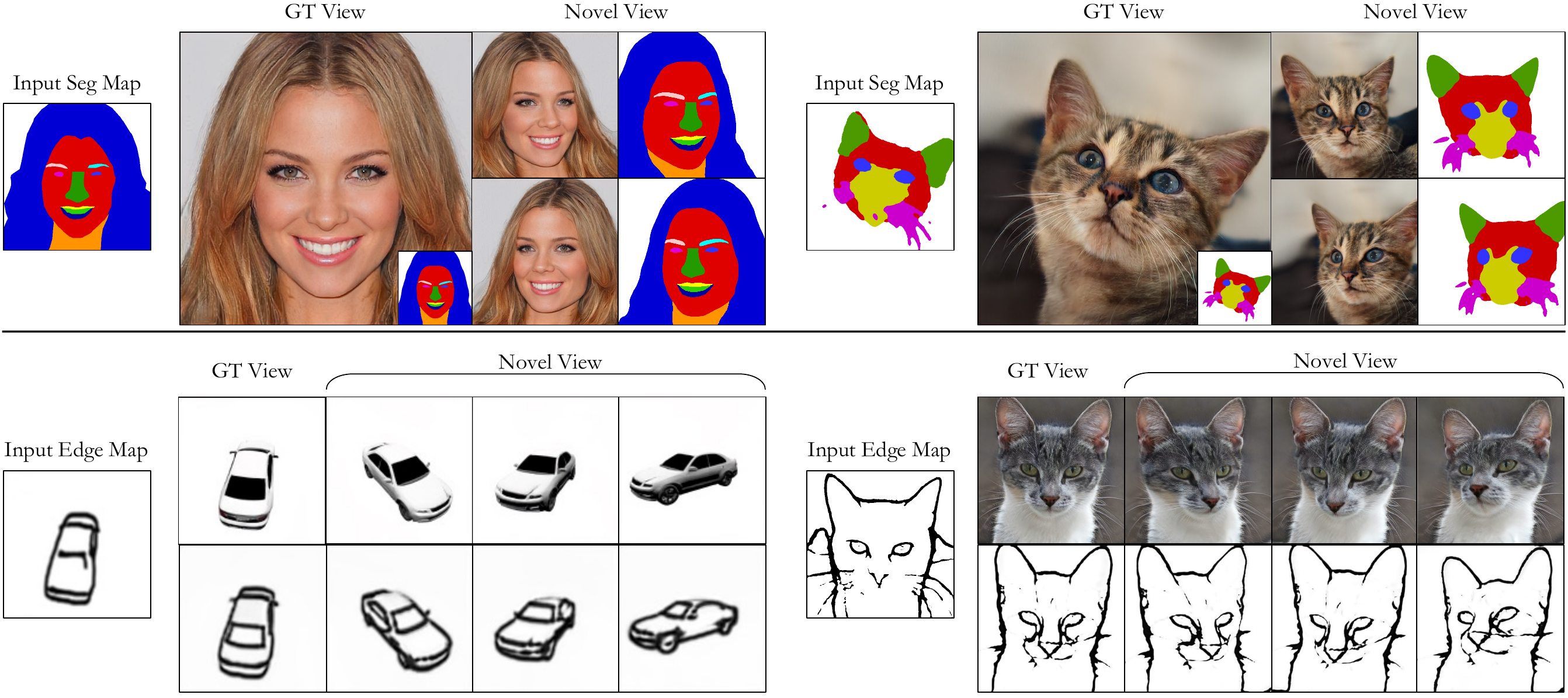}
 \vspace{-1.5em}
\captionof{figure}{Given a 2D label map as input, such as a segmentation or edge map, our model learns to predict high-quality 3D labels, geometry, and appearance, which enables us to render both labels and RGB images from different viewpoints.
The inferred 3D labels further allow interactive editing of label maps from any viewpoint, \revision{as shown in \reffig{edit_car}}. 
}
 \vspace{1em}
\lblfig{teaser}
}]

%% file: sections/00_abstract.tex
\begin{abstract}
\vspace{-5pt}
    We propose \ourmethod, a 3D-aware conditional generative model for controllable photorealistic image synthesis. Given a 2D label map, such as a segmentation or edge map, our model learns to synthesize a corresponding image from different viewpoints. 
    To enable explicit 3D user control, we extend conditional generative models with neural radiance fields. Given widely-available posed monocular image and label map pairs, our model learns to assign a label to every 3D point in addition to color and density, which enables it to render the image and pixel-aligned label map simultaneously. 
    Finally, we build an interactive system that allows users to edit the label map from different viewpoints and generate outputs accordingly. %

\end{abstract}

%% file: sections/01_introduction.tex
\section{Introduction}

Content creation with generative models has witnessed tremendous progress in recent years, enabling high-quality, user-controllable image and video synthesis~\cite{Goodfellow14,karras2019style,esser2021taming,ho2020denoising}. In particular, image-to-image translation methods~\cite{pix2pix2017,zhu2017unpaired,park2019semantic} allow users to interactively create and manipulate a high-resolution image given a 2D input label map. 
Unfortunately, existing image-to-image translation methods operate purely in 2D, without explicit reasoning of the underlying 3D structure of the content. %
As shown in \reffig{teaser}, we aim to make conditional image synthesis 3D-aware, allowing not only 3D content generation but also viewpoint manipulation and attribute editing (e.g., car shape) in 3D.

Synthesizing 3D content conditioned on user input is challenging.
For model training, it is costly to obtain large-scale datasets with paired user inputs and their desired 3D outputs. During test time, 3D content creation often requires multi-view user inputs, as a user may want to specify the details of 3D objects using 2D interfaces from different viewpoints. 
However, these inputs may not be 3D-consistent, providing conflicting signals for 3D content creation.

To address the above challenges, we extend conditional generative models with 3D neural scene representations. %
To enable \emph{cross-view} editing, we additionally encode semantic information in 3D, which can then be rendered as 2D label maps from different viewpoints. 
We learn the aforementioned 3D representation using only 2D supervision in the form of image reconstruction and adversarial losses. While the reconstruction loss ensures the alignment between 2D user inputs and corresponding 3D content, our pixel-aligned conditional discriminator encourages the appearance and labels to look plausible while remaining pixel-aligned when rendered into novel viewpoints. We also propose a cross-view consistency loss to enforce the latent codes to be consistent from different viewpoints.
\input{figTex/fig_method}

We focus on 3D-aware semantic image synthesis on the CelebAMask-HQ~\cite{CelebAMask-HQ}, AFHQ-cat\cite{choi2020starganv2}, and shapenet-car\cite{shapenet2015} datasets. Our method works well for various 2D user inputs, including segmentation maps and edge maps.
Our method outperforms several 2D and 3D baselines, such as Pix2NeRF variants~\cite{cai2022pix2nerf}, SofGAN~\cite{chen2021sofgan}, and SEAN~\cite{zhu2020sean}. %
We further ablate the impact of various design choices and demonstrate applications of our method, such as cross-view editing and explicit user control over semantics and style. Please see our \href{http://cs.cmu.edu/~pix2pix3D}{website} for more results and \href{https://github.com/dunbar12138/pix2pix3D}{code}.

%% file: figTex/fig_method.tex
\begin{figure*}[h!]
    \centering
    \includegraphics[width=\linewidth]{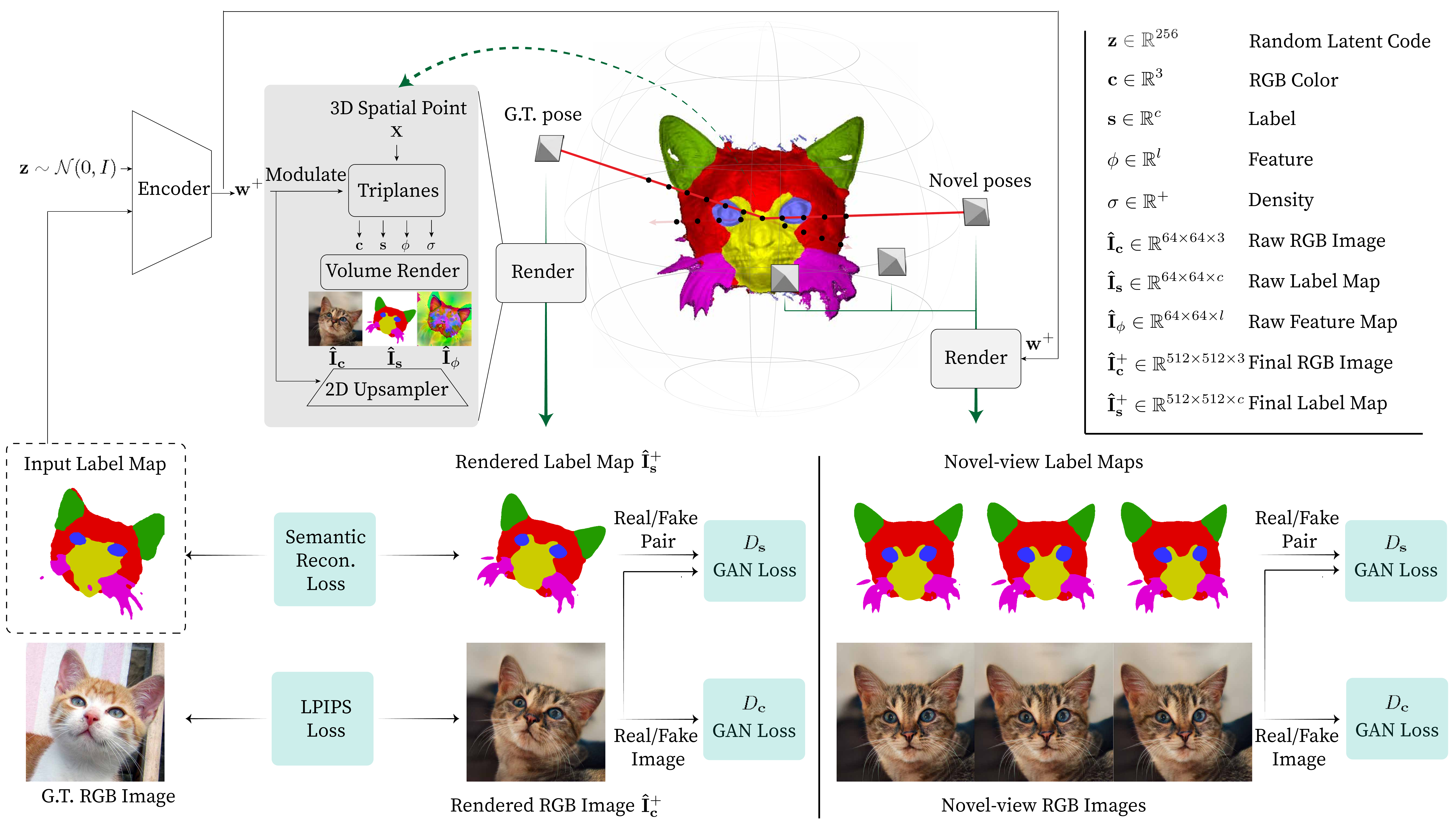}
    \vspace{-2em}
    \caption{\looseness=-1\textbf{Overall pipeline.} Given a 2D label map (e.g., segmentation map), a random latent code $z$, and a camera pose $\hat P$ as inputs, our generator renders the label map and image from viewpoint $\hat P$. Intuitively, the input label map specifies the geometric structure, while the latent code captures the appearance, such as hair color. We begin with an encoder that encodes both the input label map and the latent code into style vectors $\mathbf{w}^{+}$. We then use $\mathbf{w}^{+}$ to modulate our 3D representation, which takes a spatial point $\mathbf{x}$ and outputs (1) color $\mathbf{c}\in \mathbb{R}^3$, (2) density $\sigma$, (3) feature $\phi \in \mathbb{R}^{l}$, and (4) label $\mathbf{s} \in \mathbb{R}^{c}$. We then perform volumetric rendering and 2D upsampling to get the high-res label map $\mathbf{\hat I}_{\mathbf{s}}^+$ and RGB Image $\mathbf{\hat I}_{\mathbf{c}}^+$. For those rendered from ground-truth poses, we compare them to ground-truth labels and images with an LPIPS loss and label reconstruction loss. We apply a GAN loss on labels and images rendered from both novel and original viewpoints. %
    }
    \lblfig{method}
    \vspace{-0.5em}
\end{figure*}

%% file: sections/02_related_work.tex
\section{Related Work}
\lblsec{related_work}

\myparagraph{Neural Implicit Representation.} Neural implicit fields, such as DeepSDF and NeRFs~\cite{mildenhall2020nerf, park2019deepsdf}, model the appearance of objects and scenes with an implicitly defined, continuous 3D representation parameterized by neural networks. %
They have produced significant results for 3D reconstruction~\cite{Zhu2022CVPR,Sucar:etal:ICCV2021} and novel view synthesis applications~\cite{zhang2020nerf++, martinbrualla2020nerfw,meng2021gnerf, lin2021barf,mueller2022instant} thanks to their compactness and expressiveness. %
NeRF and its descendants aim to optimize a network for an individual scene, given hundreds of images from multiple viewpoints. Recent works further reduce the number of training views through learning network initializations~\cite{yu2020pixelnerf,tancik2021learned,chen2021mvsnerf}, leveraging auxiliary supervision~\cite{kangle2021dsnerf,jain2021putting}, or imposing regularization terms~\cite{niemeyer2021regnerf}. Recently, explicit or hybrid representations of radiance fields \cite{mueller2022instant,yu_and_fridovichkeil2021plenoxels,Chen2022ECCV} have also shown promising results regarding quality and speed. In our work, we use hybrid representations for modeling both user inputs and outputs in 3D, focusing on synthesizing novel images rather than reconstructing an existing scene. A recent work Pix2NeRF~\cite{cai2022pix2nerf} aims to translate a single image to a neural radiance field, which allows single-image novel view synthesis. In contrast, we focus on 3D-aware user-controlled content generation. %

\myparagraph{Conditional GANs.} Generative adversarial networks (GANs) learn the distribution of natural images by forcing the generated and real images to be indistinguishable. 
They have demonstrated high-quality results on 2D image synthesis and manipulation~\cite{Goodfellow14,karras2019style,karras2020analyzing, Karras2021, zhu2016generative,Brock2018Biggan,patashnik2021styleclip,zhu2020domain,tov2021designing,shen2021closed,bau2019semantic,abdal2019image2stylegan}. Several methods adopt image-conditional GANs~\cite{pix2pix2017,mirza2014conditional} for user-guided image synthesis and editing applications~\cite{zhu2017unpaired,liu2017unsupervised,wang2018pix2pixHD,zhu2020sean,park2019semantic,park2020contrastive,huang2018multimodal,huang2018,schonfeld2020you,CelebAMask-HQ} . 
In contrast, we propose a 3D-aware generative model conditioned on 2D user inputs that can render view-consistent images and enable interactive 3D editing. Recently, 
SoFGAN \cite{chen2021sofgan} uses a 3D semantic map generator and a 2D semantic-to-image generator to enable 3D-aware generation, but using 2D generators does not ensure 3D consistency.

\myparagraph{3D-aware Image Synthesis.} %
Early data-driven 3D image editing systems can achieve various 3D effects but often require a huge amount of manual effort~\cite{kholgade20143d
,chen20133}. 
Recent works have integrated the 3D structure into learning-based image generation pipelines using various geometric representations, including voxels~\cite{zhu2018visual,henzler2019escaping}, voxelized 3D features~\cite{nguyen2019hologan}, and 3D  morphable models~\cite{yao20183d,tewari2020stylerig}. However, many rely on external 3D data~\cite{zhu2018visual,tewari2020stylerig,yao20183d}.  
Recently, neural scene representations have been integrated into GANs to enable 3D-aware image synthesis~\cite{schwarz2020graf,niemeyer2021giraffe,gu2022stylenerf, EG3D, chan2021pi,or2021stylesdf,xu20213d,pan2021shading}. Intriguingly, these 3D-aware GANs can learn 3D structures without any 3D supervision. 
For example, StyleNeRF~\cite{gu2022stylenerf} and EG3D~\cite{EG3D} learn to generate 3D representations by modulating either NeRFs or explicit representations with latent style vectors. This allows them to render high-resolution view-consistent images.  %
Unlike the above methods, we focus on conditional synthesis and interactive editing rather than random sampling. \revision{Several works~\cite{igarashi2006teddy,xie2013sketch,lun20173d,delanoy2018you} have explored sketch-based shape generation but they do not allow realistic image synthesis. }%

Closely related to our work, Huang et al.~\cite{huang2020semantic} propose synthesizing novel views conditional on a semantic map. Our work differs in three ways. First, we can predict full 3D labels, geometry, and appearance, rather than only 2D views, which enables cross-view editing.
Second, our method can synthesize images with a much wider baseline than Huang et al. ~\cite{huang2020semantic}. Finally, our learning algorithm does not require ground truth multi-view images of the same scene. Two recent works, FENeRF~\cite{sun2021fenerf} and 3DSGAN~\cite{zhang20213d}, also leverage semantic labels for training 3D-aware GANs, but they do not support conditional inputs and require additional efforts (e.g., GAN-inversion) to allow user editing.
\revision{Three concurrent works, IDE-3D~\cite{sun2022ide}, NeRFFaceEditing~\cite{jiang2022nerffaceediting}, and sem2nerf~\cite{chen2022sem2nerf}, also explore the task of 3D-aware generation based on segmentation masks. However, IDE-3D and sem2nerf only allow editing on a fixed view, and NeRFFaceEditing focuses on real image editing rather than generation. All of them do not include results for other input modalities. In contrast, we present a general-purpose method that works well for diverse datasets and input controls.}

%% file: sections/03_method.tex
\section{Method}

Given a 2D label map ${\mathbf{I}}_\mathbf{s}$, such as a segmentation or edge map, \ourmethod generates a 3D-volumetric representation of geometry, appearance, and labels that can be rendered from different viewpoints. \reffig{method} provides an overview. 
We first introduce the formulation of our 3D conditional generative model for 3D-aware image synthesis in \refsec{representations}. 
Then, in \refsec{training}, we discuss how to learn the model from color and label map pairs $\{\mathbf{I}_{\mathbf{c}}, \mathbf{I}_{\mathbf{s}}\}$ associated with poses $\mathbf{P}$.

\subsection{Conditional 3D Generative Models}
\lblsec{representations}

Similar to EG3D~\cite{EG3D}, we adopt a hybrid representation for the density and appearance of a scene and use style vectors to modulate the 3D generations. %
To condition the 3D representations on 2D label map inputs, we introduce a conditional encoder that maps a 2D label map into a latent style vector. 
Additionally, \ourmethod  produces 3D labels that can be rendered from different viewpoints, allowing for cross-view user editing.

\myparagraph{Conditional Encoder.} Given a 2D label map input ${\mathbf{I}}_\mathbf{s}$ and a random latent code sampled from the spherical Gaussian space $\mathbf{z} \sim \mathcal{N}(0,I)$, our conditional encoder $E$ 
outputs a list of style vectors $\mathbf{w}^{+} \in \mathbb{R}^{l \times 256}$,
\[
\mathbf{w}^{+} = E({\mathbf{I}}_\mathbf{s}, \mathbf{z}),
\]
where $l=13$ is the number of layers to be modulated.
Specifically, we encode ${\mathbf{I}}_\mathbf{s}$ into the first $7$ style vectors that represent the global geometric information of the scene. We then feed the random latent code $\mathbf{z}$ through a Multi-Layer Perceptron (MLP) mapping network to obtain the rest of the style vectors that control the appearance. 

\myparagraph{Conditional 3D Representation.} Our 3D representation is parameterized by tri-planes followed by an $2$-layer MLP $f$ \cite{EG3D}, which takes in a spatial point $\mathbf{x}\in \mathbb{R}^3$ and returns 4 types of outputs: (1) color $\mathbf{c}\in \mathbb{R}^3$, %
(2) density $\sigma \in \mathbb{R}^+$, (3) feature $\phi \in \mathbb{R}^{64}$ for the purpose of 2D upsampling, and most notably, (4) label $\mathbf{s} \in \mathbb{R}^{c}$, where $c$ is the number of classes if ${\mathbf{I}}_\mathbf{s}$ is a segmentation map, otherwise 1 for edge labels. We make the field conditional by modulating the generation of tri-planes $F^{\text{tri}}$ with the style vectors $\mathbf{w}^{+}$. We also remove the view dependence of the color following \cite{EG3D, gu2022stylenerf}. Formally,
\begin{equation*}
    (\mathbf{c}, \mathbf{s}, \sigma, \phi) = f(F^{\text{tri}}_{\mathbf{w}^{+}}(\mathbf{x})).
\end{equation*}

\myparagraph{Volume Rendering and Upsampling.} We apply volumetric rendering to synthesize color images~\cite{kajiya1984ray,mildenhall2020nerf}. In addition, we render label maps, which are crucial for enabling cross-view editing (\refsec{application}) and improving rendering quality (\reftbl{seg2face}). Given a viewpoint $\hat P$ looking at the scene origin, we sample $N$ points along the ray that emanates from a pixel location and query density, color, labels, and feature information from our 3D representation. 
Let ${\bf x_i}$ be the i-th sampled point along the ray $r$.
Let ${\bf c}_i, {\bf s}_i$ and ${\bf \phi}_i$ be the color, labels, and the features of ${\bf x_i}$.
Similar to \cite{sun2021fenerf}, 
The color, label map, and feature images are computed as the weighted combination of queried values,
\begin{equation}
\lbleq{volume_render}
    {\bf \hat I_c}(r) = \sum_{i=1}^{N} \tau_i{\bf c}_i, \;\; {\bf \hat I_s}(r) = \sum_{i=1}^{N} \tau_i{\bf s}_i , \;\;  {\bf \hat I_{\phi}}(r) = \sum_{i=1}^{N} \tau_i{\bf {\phi}}_i,
\end{equation}
where the transmittance ${\tau_i}$ is computed as the probability of a photon traversing between the camera center and the i-th point given the length of the i-th interval $\delta_i$,
\begin{equation*}
    {\tau_i} = \prod_{j=1}^{i} \exp{(-\sigma_j\delta_j)} (1-\exp{(-\sigma_i\delta_i})).
\end{equation*}
Similar to prior works \cite{gu2022stylenerf,or2021stylesdf,EG3D}, we approximate \refeq{volume_render} by 2D Upsampler $U$ to reduce the computational cost. 
We render high-res $512\times 512$ images in two passes. In the first pass, we render low-res $64\times 64$ images ${\bf \hat I_{c}}, {\bf \hat I_{s}}, {\bf \hat I_{\phi}}$. Then a CNN up-sampler $U$ is applied to obtain high-res images,
\begin{equation*}
    \mathbf{\hat I}_{\mathbf{c}}^+ = U({\bf \hat I_{c}},{\bf \hat I_{\phi}}), \qquad \mathbf{\hat I}_{\mathbf{s}}^+ =  U({\bf \hat I_{s}}, {\bf \hat I_{\phi}}).
\end{equation*}

\subsection{Learning Objective}
\lblsec{training}

Learning conditional 3D representations from monocular images is challenging due to its under-constrained nature. 
Given training data of associated images, label maps, and camera poses predicted by an off-the-shelf model, we carefully construct learning objectives, including reconstruction, adversarial, and cross-view consistency losses. These objectives will be described below. %

\myparagraph{Reconstruction Loss.} 
Given a ground-truth viewpoint ${\bf  P}$ associated with the color and label maps $\{{\mathbf{I}}_{\mathbf{c}}, {\mathbf{I}}_{\mathbf{s}}\}$, we render color and label maps from ${\bf  P}$ and compute reconstruction losses for both high-res and low-res output.
We use LPIPS~\cite{zhang2018perceptual} to compute the image reconstruction loss $\mathcal{L}_c$ for color images. For label reconstruction loss $\mathcal{L}_s$, we use the balanced cross-entropy loss for segmentation maps or L2 Loss for edge maps, 
\begin{equation*}
    \mathcal{L}_{\text{recon}} = \lambda_c \mathcal{L}_c({\mathbf{I}}_{\mathbf{c}},\{\mathbf{\hat I}_{\mathbf{c}}, \mathbf{\hat I}_{\mathbf{c}}^+\})+ \lambda_s \mathcal{L}_s({\mathbf{I}}_{\mathbf{s}}, \{\mathbf{\hat I}_{\mathbf{s}}, \mathbf{\hat I}_{\mathbf{s}}^+\}),
\end{equation*}
where $\lambda_c$ and $\lambda_s$ balance two terms. 

\myparagraph{Pixel-aligned Conditional Discriminator.} 
The reconstruction loss alone fails to synthesize detailed results from novel viewpoints. Therefore, we use an adversarial loss~\cite{Goodfellow14} to enforce renderings to look realistic from random viewpoints. 
Specifically, we have two discriminators $D_{\mathbf{c}}$ and $D_{\mathbf{s}}$ for RGB images and label maps, respectively. $D_{\mathbf{c}}$ is a widely-used GAN loss that takes real and fake images as input, while the pixel-aligned conditional discriminator  $D_{\mathbf{s}}$ concatenates color images and label maps as input, which encourages pixel alignment between color images and label maps. 
Notably, in $D_{\mathbf{s}}$, we stop the gradients for the color images to prevent a potential quality downgrade. We also feed the rendered low-res images to prevent the upsampler from hallucinating details, inconsistent with the low-res output. The adversarial loss can be written as follows. 
\begin{equation*}
    \mathcal{L}_{\text{GAN}} = \lambda_{D_{\bf c}}\mathcal{L}_{D_{\mathbf{c}}}(\mathbf{\hat I}_{\mathbf{c}}^+, \mathbf{\hat I}_{\mathbf{c}}) + \lambda_{D_{\bf s}} \mathcal{L}_{D_{\mathbf{s}}}(\mathbf{\hat I}_{\mathbf{c}}^+, \mathbf{\hat I}_{\mathbf{c}}, \mathbf{\hat I}_{\mathbf{s}}^+, \mathbf{\hat I}_{\mathbf{s}}).
\end{equation*}
where $\lambda_{D_{\bf c}}$ and $\lambda_{D_{\bf s}}$ balance two terms.
To stabilize the GAN training, we adopt the R1 regularization loss~\cite{mescheder2018training}.

\myparagraph{Cross-view Consistency Loss.} We observe that inputting label maps of the same object from different viewpoints will sometimes result in different 3D shapes. Therefore we add a cross-view consistency loss to regularize the training, as illustrated in \reffig{CVC_loss}. Given an input label map ${\mathbf{I}}_{\mathbf{s}}$ and its associated pose ${\bf  P}$, we generate the label map $\mathbf{\hat I}_{\mathbf{s}}'$ from a different viewpoint ${\bf P}'$, and render the label map $\mathbf{\hat I}_{\mathbf{s}}''$ back to the pose ${\bf P}$  using $\mathbf{\hat I}_{\mathbf{s}}'$ as input. We add a reconstruction loss between $\mathbf{\hat I}_{\mathbf{s}}''$ and $\mathbf{\hat I}_{\mathbf{s}}$:
\begin{equation*}
    \mathcal{L}_{\text{CVC}} = \lambda_{\text{CVC}}\mathcal{L}_s(\mathbf{\hat I}_{\mathbf{s}}'',\mathbf{\hat I}_{\mathbf{s}}),
\end{equation*}
where $\mathcal{L}_s$ denotes the reconstruction loss in the label space, and $\lambda_{\text{CVC}}$ weights the loss term. This loss is crucial for reducing error accumulation during cross-view editing.

\input{figTex/fig_CVC.tex}
\myparagraph{Optimization.} Our final learning objective is written as follows:
\begin{equation*}
    \mathcal{L}_{\text{total}} =  \mathcal{L}_{\text{recon}} + \mathcal{L}_{\text{GAN}} + \mathcal{L}_{\text{CVC}}. 
\end{equation*}
At every iteration, we determine whether to use a ground-truth pose or sample a random one with a probability of $p$. 
We use the reconstruction loss and GAN loss for ground-truth poses, while for random poses, we only use the GAN loss. 
We provide the hyper-parameters and more implementation details in \refapp{details}.

\input{figTex/fig_baselines.tex}

\input{tabTex/tab_seg2face.tex}
\input{tabTex/tab_shapenet_v1.tex}

\input{figTex/fig_expA}

\input{figTex/fig_edge2cat.tex}
\input{figTex/fig_edge2car.tex}

\input{tabTex/tab_AFHQ_seg.tex}

%% file: figTex/fig_CVC.tex
\begin{figure}[!t]
    \centering
    \includegraphics[width=\linewidth]{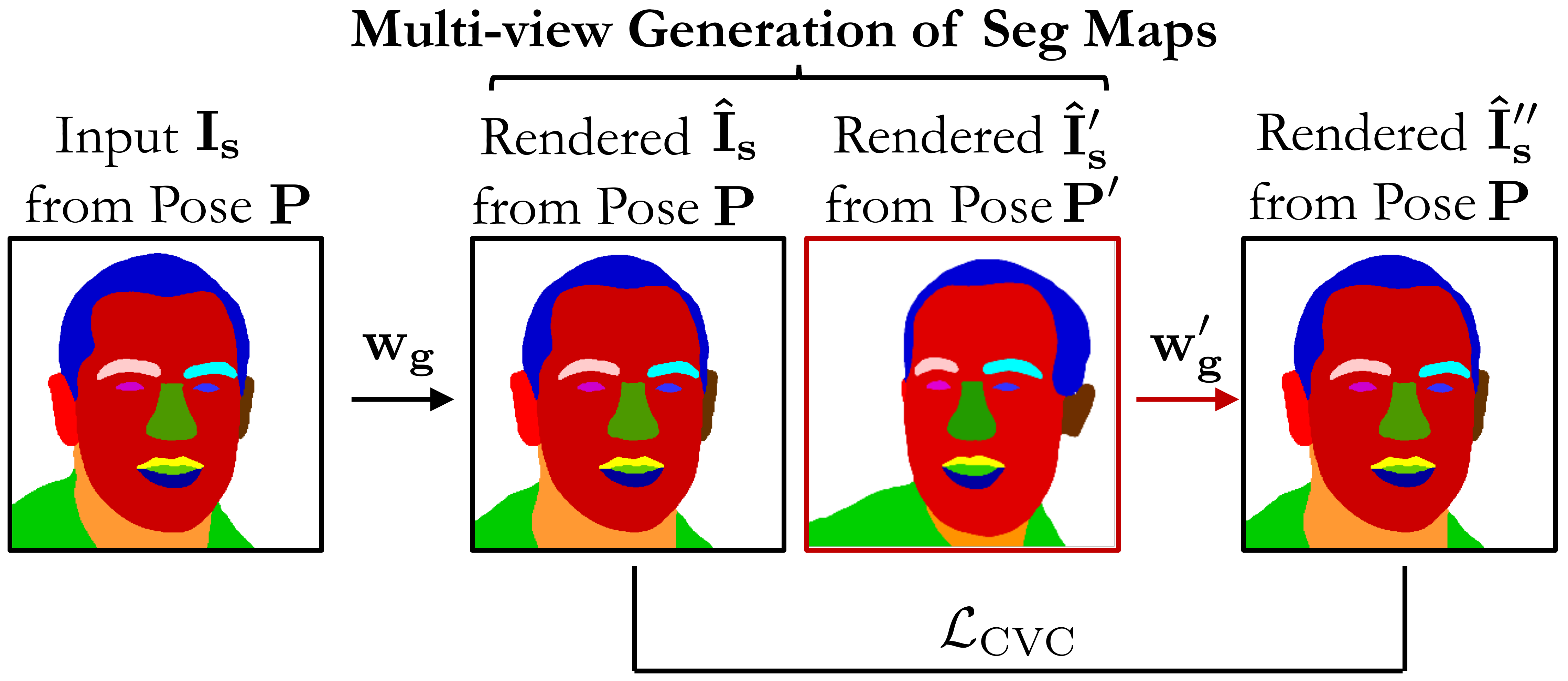}
    \caption{\textbf{Cross-View Consistency Loss.} Given an input label map ${\mathbf{I}}_{\mathbf{s}}$ and its associated pose ${\bf P}$, we first infer the geometry latent code ${{\bf w_g}}$. From ${{\bf w_g}}$, we can generate a label map ${\mathbf{\hat I}}_{\mathbf{s}}$ from the same pose ${\bf P}$, and $\mathbf{\hat I}_{\mathbf{s}}'$ from a random pose ${\bf P'}$. Next, we infer ${\bf w_g'}$ from the novel view $\mathbf{\hat I}_{\mathbf{s}}'$, and render it back to the original pose ${\bf P}$ to obtain $\mathbf{\hat I}_{\mathbf{s}}''$. Finally, we add a reconstruction loss:  
    $\mathcal{L}_{\text{CVC}} = \lambda_{\text{CVC}}\mathcal{L}_s(\mathbf{\hat I}_{\mathbf{s}}'',\mathbf{\hat I}_{\mathbf{s}})$.
    }
    \lblfig{CVC_loss}
\end{figure}

%% file: figTex/fig_baselines.tex
\begin{figure*}[t]
\centering
\includegraphics[width=\textwidth]{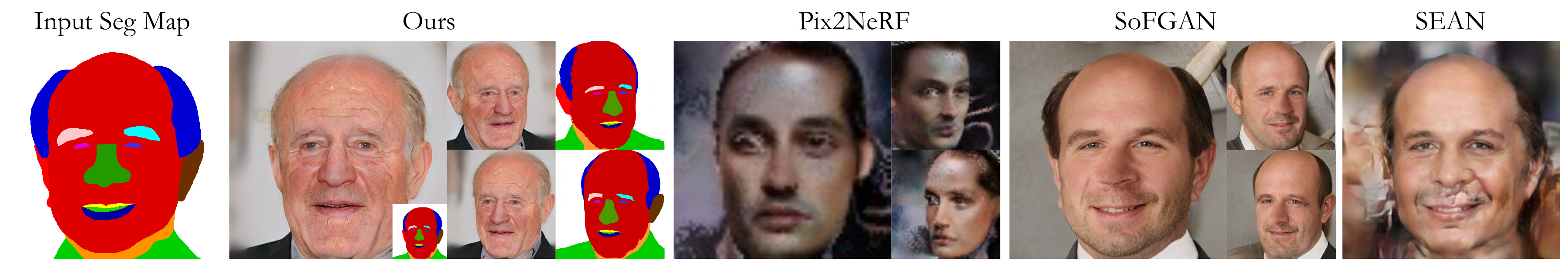}
 \vspace{-2em}
\captionof{figure}{
\textbf{Qualitative Comparison with Pix2NeRF\cite{cai2022pix2nerf}, SoFGAN\cite{chen2021sofgan}, and SEAN\cite{zhu2020sean}} on CelebAMask dataset for seg2face task. SEAN fails in multi-view synthesis, while SoFGAN suffers from multi-view inconsistency (e.g., face identity changes across viewpoints). Our method renders high-quality images while maintaining multi-view consistency. Please check our \href{http://cs.cmu.edu/~pix2pix3D}{website} for more examples.
}

\vspace{-1em}
\lblfig{seg2face_baselines}
\end{figure*}

%% file: tabTex/tab_seg2face.tex
\begin{table}[t]
\setlength{\tabcolsep}{2pt}
\setlength{\extrarowheight}{4pt}
\renewcommand{\arraystretch}{0.8}
\centering
\scriptsize
\begin{tabular}{lccc|cc|c}
\toprule
\multirow{2}{*}{Seg2Face}  & \multicolumn{3}{c}{\textsc{Quality}} & \multicolumn{2}{c}{\textsc{Alignment}} \\ %
& & & SG & & & FVV\\
\textsc{CelebAMask\cite{CelebAMask-HQ}} & FID $\downarrow$  & KID $\downarrow$ & Diversity $\uparrow$  & mIoU $\uparrow$ & acc $\uparrow$ & Identity $\downarrow$ \\ \midrule
\textsc{SEAN \cite{zhu2020sean}} & 32.74 & 0.018 & 0.29 & 0.52 & 0.85 & N/A \\
\textsc{SoFGAN \cite{chen2021sofgan}} & 23.34 & 0.012 & \textbf{0.33} & \textbf{0.53} & 0.89 & 0.58   \\
\textsc{Pix2NeRF \cite{cai2022pix2nerf}} & 54.23 & 0.042 & 0.16 & 0.36 & 0.65 & 0.44 \\
\midrule
\textsc{\textbf{\ourmethod{} (Ours})} \\
\textsc{w/o 3D Labels} & 12.96 & 0.005 & 0.30 & N/A (0.43)  & N/A (0.81) & 0.38  \\
\textsc{w/o CVC} & 11.62 & 0.004 & 0.30 & 0.50 (0.50)  & 0.87 (0.85) & 0.42 \\
\textsc{full model} & 11.54 & \textbf{0.003} &  0.28 & 0.51 (0.52) & \textbf{0.90} (0.88) & \textbf{0.36} \\
\textsc{full model}$^\dagger$  & \textbf{11.13} & \textbf{0.003} &  0.29 & 0.51 (0.50) & \textbf{0.90} (0.87) & \textbf{0.36} \\

\bottomrule
\end{tabular}
\caption{\textbf{Seg2face Evaluation.} 
Our metrics include image quality (FID, KID, SG Diversity), alignment (mIoU and acc against GT label maps), and multi-view consistency (FVV Identity). Single-generation diversity (SG Diversity) is obtained by computing the LPIPS metric between randomly generated pairs given a single conditional input. To evaluate alignment, we compare the generated label maps against the ground truth in terms of mIoU and pixel accuracy (acc). Alternatively, given a generated image, one could estimate label maps via a face parser, and compare those against the ground truth (numbers in parentheses).
We include SEAN \cite{zhu2020sean} and SoFGAN \cite{chen2021sofgan} as baselines, and modify Pix2NeRF \cite{cai2022pix2nerf} to take conditional input. Our method achieves the best quality, alignment ACC, and FVV Identity while being competitive on SG Diversity. SoFGAN tends to have better alignment but worse 3D consistency. We also ablate our method w.r.t the 3D labels and the cross-view consistency (CVC) loss. 
Our 3D labels are crucial for alignment, while the CVC loss improves multi-view consistency. 
Using pretrained models from EG3D ($\dagger$) also improves the performance.
}
\vspace{-1em}

\lbltbl{seg2face}
\end{table}

%% file: tabTex/tab_shapenet_v1.tex
\begin{table}[t]
\setlength{\tabcolsep}{4pt}
\setlength{\extrarowheight}{5pt}
\renewcommand{\arraystretch}{0.8}
\centering
\scriptsize
\begin{tabular}{lccc|c}
\toprule
\multirow{2}{*}{Edge2Car } & \multicolumn{3}{c}{\textsc{Quality}} & \multicolumn{1}{c}{\textsc{Alignment}} \\ %
               & FID $\downarrow$  & KID $\downarrow$ & SG Diversity $\uparrow$ & AP $\uparrow$   \\ \midrule
\textsc{Pix2NeRF \cite{cai2022pix2nerf}} & 23.42 & 0.014 & 0.06 & 0.28 \\

\midrule
\textsc{\textbf{\ourmethod{} (Ours)}} \\
\textsc{w/o 3D Labels} & 10.73 & 0.005 & 0.12 & 0.45 (0.42) \\
\textsc{w/o CVC} & 9.42 & \textbf{0.004} & \textbf{0.13} & 0.61 (0.59) \\
\textsc{full model} & \textbf{8.31} & \textbf{0.004} & \textbf{0.13} & \textbf{0.63} (0.59) \\

\bottomrule
\end{tabular}
\caption{\textbf{Edge2car Evaluation.} We compare our method with Pix2NeRF\cite{cai2022pix2nerf} on edge2car using the shapenet-car\cite{shapenet2015} dataset. Similar to \reftbl{seg2face}, we evaluate FID, KID, and SG Diversity for image quality. We also evaluate the alignment with the input edge map using AP. Similarly, we can either run informative drawing\cite{chan2022drawings} on generated images to obtain edge maps (numbers in parentheses) or directly use generated edge maps to calculate the metrics. We achieve better image quality and alignment than Pix2NeRF. We also find that using 3D labels and cross-view consistency loss is helpful regarding FID and AP metrics.}

\vspace{-1em}
\lbltbl{edge2car}
\end{table}

%% file: figTex/fig_expA.tex
\begin{figure*}[t]
\centering
\includegraphics[width=\linewidth]{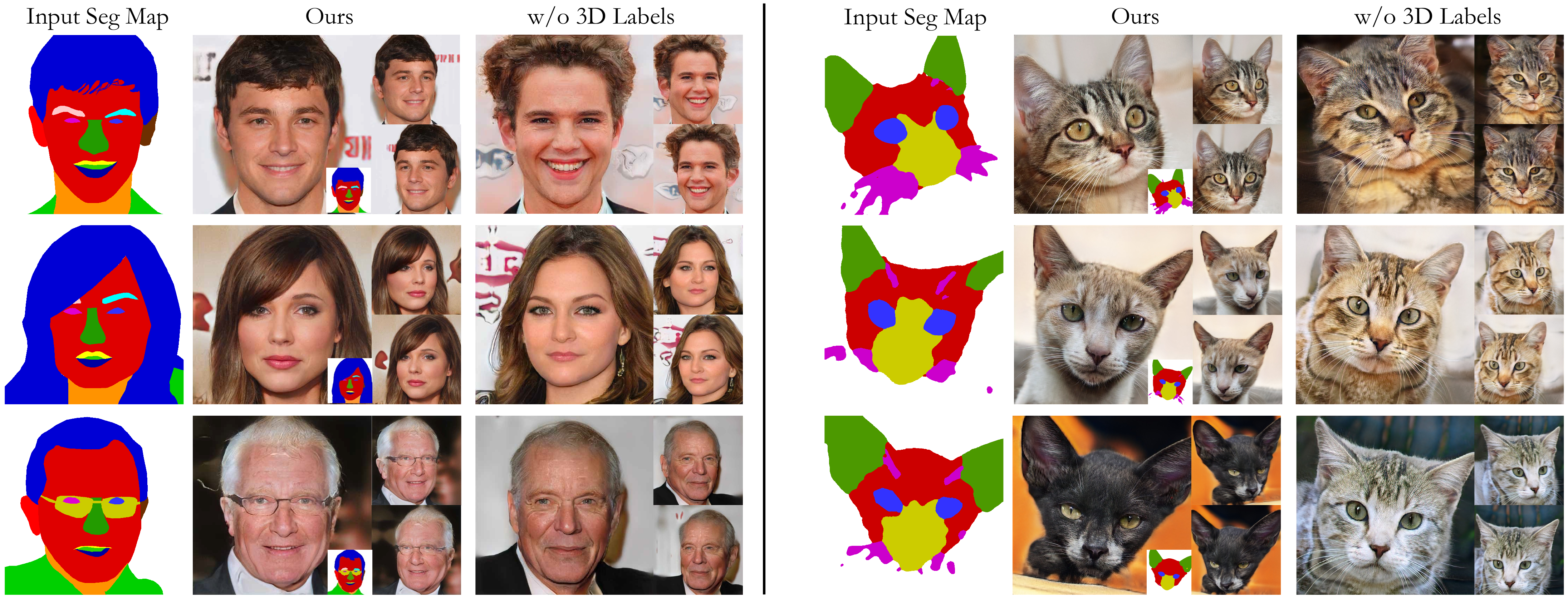}
 \vspace{-2em}
\captionof{figure}{\textbf{Qualitative ablation on seg2face and seg2cat.} We ablate our method by removing the branch that renders label maps (\textit{w/o 3D Labels}). Our results better align with input labels (e.g., hairlines and the cat's ear).
}

\vspace{-1em}
\lblfig{expA}
\end{figure*}

%% file: figTex/fig_edge2cat.tex
\begin{figure}[t]
    \centering
    
    \includegraphics[width=\linewidth]{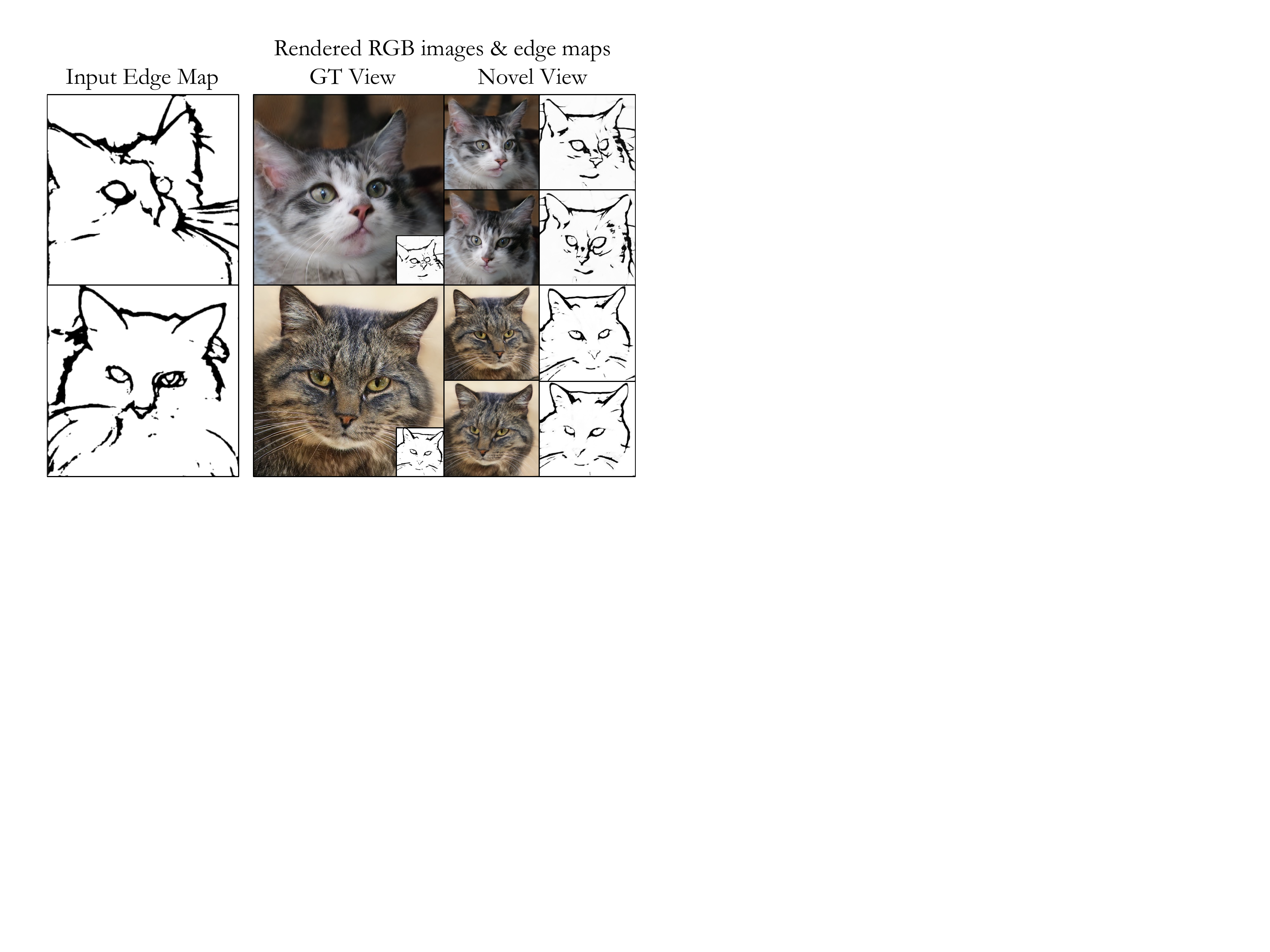}
    \vspace{-2em}
    \caption{\textbf{Results on edge2cat.} Our model is trained on AFHQ-cat\cite{choi2020starganv2} with edges extracted by \revision{pidinet}\cite{su2021pdc}.} 
    \lblfig{edge2cat}
    \vspace{-1em}
\end{figure}

%% file: figTex/fig_edge2car.tex
\begin{figure}[t]
    \centering
    \includegraphics[width=\linewidth]{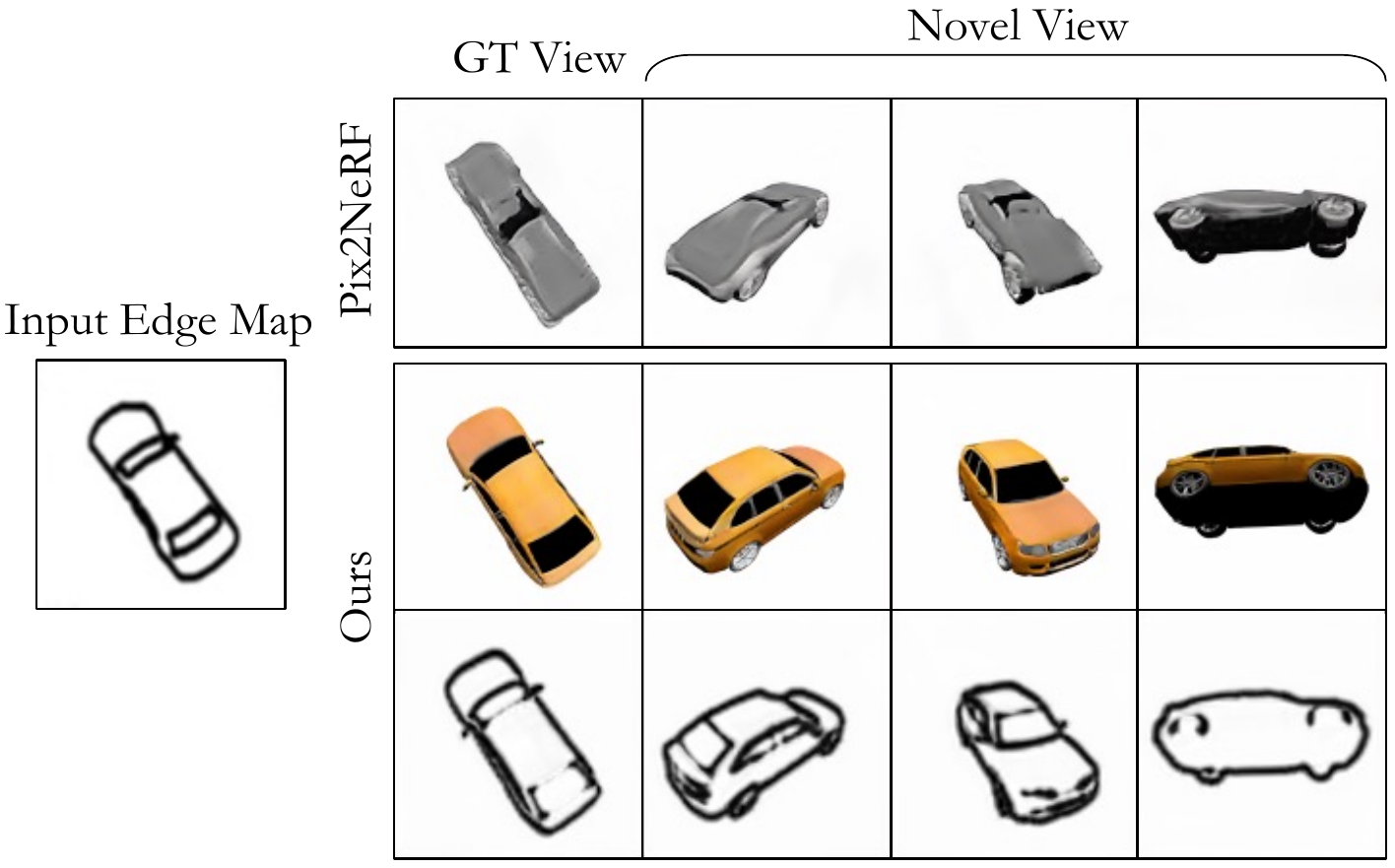}
    \vspace{-2em}
    \caption{\textbf{Qualitative comparisons on edge2car.} \ourmethod{} (Ours) and Pix2NeRF\cite{cai2022pix2nerf} are trained on shapenet-car\cite{shapenet2015}, and \ourmethod{} achieves better quality and alignment than Pix2NeRF.} 
    \lblfig{edge2car}
    \vspace{-1em}
\end{figure}

%% file: tabTex/tab_AFHQ_seg.tex
\begin{table}[t]
\setlength{\tabcolsep}{4pt}
\setlength{\extrarowheight}{5pt}
\renewcommand{\arraystretch}{0.8}
\scriptsize
\centering
\begin{tabular}{lccc|cc}
\toprule
Seg2Cat & \multicolumn{3}{c}{\textsc{Quality}} & \multicolumn{2}{c}{\textsc{Alignment}} \\ %
  \textsc{AFHQ-cat}\cite{karras2019style}              & FID $\downarrow$  & KID $\downarrow$ & SG Diversity $\uparrow$ & mIoU $\uparrow$ & acc $\uparrow$  \\ \midrule
\textsc{Pix2NeRF \cite{cai2022pix2nerf}} & 43.92 & 0.081 & 0.15 & 0.27 & 0.58 \\

\midrule
\textsc{\textbf{Ours}} \\
\textsc{w/o 3D Labels} & 10.41 & 0.004 & 0.26 & N/A (0.49) & N/A (0.69) \\
\textsc{w/o CVC} & 9.64 & 0.004 & 0.26 & \textbf{0.66} (0.63) & 0.76 (0.73) \\
\textsc{Full Model} & \textbf{8.62} & \textbf{0.003} & \textbf{0.27} & \textbf{0.66} (0.62) & \textbf{0.78} (0.73)  \\

\bottomrule
\end{tabular}
\caption{\textbf{Seg2cat Evaluation.} We compare our method with Pix2NeRF \cite{cai2022pix2nerf} on Seg2Cat using AFHQ-cat dataset\cite{choi2020starganv2}, with segmentation obtained by clustering DINO features \cite{amir2021deep}. Similar to \reftbl{seg2face}, we evaluate the image quality and alignment. Ours performs better in all metrics.}  %
\vspace{-1em}
\lbltbl{baseline_app}
\end{table}

%% file: sections/04_experiment.tex
\input{figTex/fig_mesh}
\input{figTex/fig_sample}

\input{figTex/fig_edit_car.tex}
\section{Experiment}

We first introduce the datasets and evaluation metrics. Then we compare our method with the baselines. Finally, we demonstrate cross-view editing and multi-modal synthesis applications enabled by our method.

\myparagraph{Datasets.}
We consider four tasks: \textit{seg2face}, \textit{seg2cat}, \textit{edge2cat}, and \textit{edge2car} in our experiments.
For seg2face, we use CelebAMask-HQ~\cite{CelebAMask-HQ} for evaluation.
CelebAMask-HQ contains 30,000 high-resolution face images from CelebA \cite{liu2015faceattributes}, and each image has a facial part segmentation mask and a predicted pose. 
The segmentation masks contain 19 classes, including skin, eyebrows, ears, mouth, lip, etc. The pose associated with each image segmentation is predicted by HopeNet~\cite{Ruiz_2018_CVPR_Workshops}.
We split the CelebAMask-HQ dataset into a training set of 24,183, a validation set of 2,993, and a test set of 2,824, following the original work~\cite{CelebAMask-HQ}.
For seg2cat and edge2cat, we use AFHQ-cat ~\cite{choi2020starganv2}, which contains 5,065 images at $512\times$ resolution. We estimate the viewpoints using unsup3d~\cite{Wu_2020_CVPR}. We extract the edges using \revision{pidinet}~\cite{su2021pdc} and obtain segmentation by clustering DINO features~\cite{amir2021deep} into 6 classes.
For edge2car, we use 3D models from shapenet-car ~\cite{shapenet2015} and render 500,000 images at $128\times$ resolution for training, and 30,000 for evaluation. We extract the edges using informative drawing~\cite{chan2022drawings}.
We train our model at $512\times$ resolution except for $128\times$ in the edge2car task. %

\myparagraph{Running Time.} %
For training the model at $512\times$ resolution, it takes about three days on eight RTX 3090 GPUs. But we can significantly reduce the training time to 4 hours if we initialize parts of our model with pretrained weights from EG3D\cite{EG3D}. During inference, our model takes 10 ms to obtain the style vector, and another 30 ms to render the final image and the label map on a single RTX A5000. The low latency (25 FPS) allows for interactive user editing. 

\subsection{Evaluation metrics}
\lblsec{evaluation}
We evaluate the models from two aspects: 1) the image quality regarding fidelity and diversity, and 2) the alignment between input label maps and generated outputs. 

\myparagraph{Quality Metrics.} Following prior works~\cite{park2019SPADE,gu2022stylenerf}, we use the \texttt{clean-fid} library~\cite{parmar2021cleanfid} to compute Fr\'echet Inception Distance (FID)~\cite{heusel2017gans} and Kernel Inception Distance (KID)~\cite{binkowski2018demystifying} to measure the distribution distance between synthesized results and real images. %
We also evaluate the single-generation diversity (SG Diversity) by calculating the LPIPS metric between randomly generated pairs given a single input following prior works~\cite{zhu2017toward,chen2021sofgan}. For FID and KID, we generate 10 images per label map in the test set using randomly sampled $z$. We compare our generated images with the whole dataset, including training and test images.

\myparagraph{Alignment Metrics.} We evaluate models on the test set using mean Intersection-over-Union (mIoU) and pixel accuracy (acc) for segmentation maps following existing works~\cite{park2019SPADE,onfeld2021you}, and average precision (AP) for edge maps. 
For those models that render label maps as output, we directly compare them with ground-truth labels. 
Otherwise, we first predict the label maps from the output RGB images using off-the-shelf networks~\cite{CelebAMask-HQ,su2021pdc}, and then compare the prediction with the ground truth. 
The metrics regarding such predicted semantic maps are reported within brackets in \reftbl{seg2face} and \reftbl{edge2car}.

For seg2face, we evaluate the preservation of facial identity from different viewpoints (FVV Identity) by calculating their distances with the dlib face recognition algorithm\footnote{\url{https://github.com/ageitgey/face_recognition}}.

\subsection{Baseline comparison}
\lblsec{comparison}

\looseness=-1
\myparagraph{Baselines.} Since there are no prior works on conditional 3D-aware image synthesis, we make minimum modifications to Pix2NeRF~\cite{cai2022pix2nerf} to be conditional on label maps instead of images. For a thorough comparison, we introduce several baselines: SEAN \cite{zhu2020sean} and SoFGAN \cite{chen2021sofgan}.  2D baselines like SEAN \cite{zhu2020sean} cannot generate multi-view images by design (N/A for FVV Identity), while SoFGAN \cite{chen2021sofgan} uses an unconditional 3D semantic map generator before the 2D generator so we can evaluate FVV Identity for that.

\myparagraph{Results.}
\reffig{seg2face_baselines} shows the qualitative comparison for seg2face and \reftbl{seg2face} reports the evaluation results. SoFGAN\cite{chen2021sofgan} tends to produce results with slightly better alignment but worse 3D consistency for its 2D RGB generator. Our method achieves the best quality, alignment acc, and FVV Identity while being competitive with 2D baselines on SG diversity. \reffig{expA} shows the qualitative ablation on seg2face and seg2cat. \reftbl{seg2car} reports the metrics for seg2cat.
\reffig{edge2cat} shows the example results for edge2cat. \reffig{edge2car} shows the qualitative comparison for edge2car and \reftbl{edge2car} reports the metrics. Our method achieves the best image quality and alignment. \reffig{mesh} shows semantic meshes of human and cat faces, extracted by marching cubes and colored by our learned 3D labels. %
We provide more evaluation results in \refapp{additional_experiments}.

\input{figTex/fig_stylemix}
\input{figTex/fig_interp}

\myparagraph{Ablation Study.} %
We compare our full method to several variants. 
Specifically, (1) \textsc{w/o 3D Labels}, we remove the branch of rendering label maps from our method, and  (2) \textsc{w/o CVC}, we remove the cross-view consistency loss. 
From \reftbl{seg2face}, \reftbl{edge2car}, and \reffig{expA}, rendering label maps is crucial for the alignment with the input. We posit that the joint learning of appearance, geometry, and label information poses strong constraints on correspondence between the input label maps and the 3D representation. Thus our method can synthesize images pixel-aligned with the inputs. Our CVC loss helps preserve the facial identity from different viewpoints.

\myparagraph{Analysis on random sampling of poses.} We study the effect of the different probabilities of sampling random poses during training, as shown in \reffig{sample}.
When sampling no random poses ($p=0$), the model best aligns with input label maps with suboptimal image quality. %
Conversely, \textit{only} sampling random poses ($p=1$) gives the best image quality but suffers huge misalignment with input label maps. 
We find $p=0.5$ achieves the balance between the image quality and the alignment with the input.

\subsection{Applications}
\lblsec{application}

\myparagraph{Cross-view Editing.} As shown in \reffig{edit_car}, our 3D editing system allows users to generate and edit label maps from any viewpoint instead of only the input view. %
The edited label map is further fed into the conditional encoder to update the 3D representation. Unlike GAN inversion~\cite{zhu2016generative}, our feed-forward conditional encoder allows fast inference of the latent code. Thus, a single forward pass of our full model takes only 40 ms on a single RTX A5000.

\myparagraph{Multi-modal synthesis and interpolation.} Like other style-based generative models \cite{karras2019style, gu2022stylenerf, Karras2019stylegan2, EG3D}, our method can disentangle the geometry and appearance information. Specifically, the input label map captures the geometry information while the randomly sampled latent code controls the appearance.
We show style manipulation results in \reffig{stylemix}. 
We can also interpolate both the geometry styles and the appearance styles (\reffig{interp}). These results show the clear disentanglement of our 3D representation. 

%% file: figTex/fig_mesh.tex
\begin{figure}
    \centering
    \includegraphics[width=\linewidth]{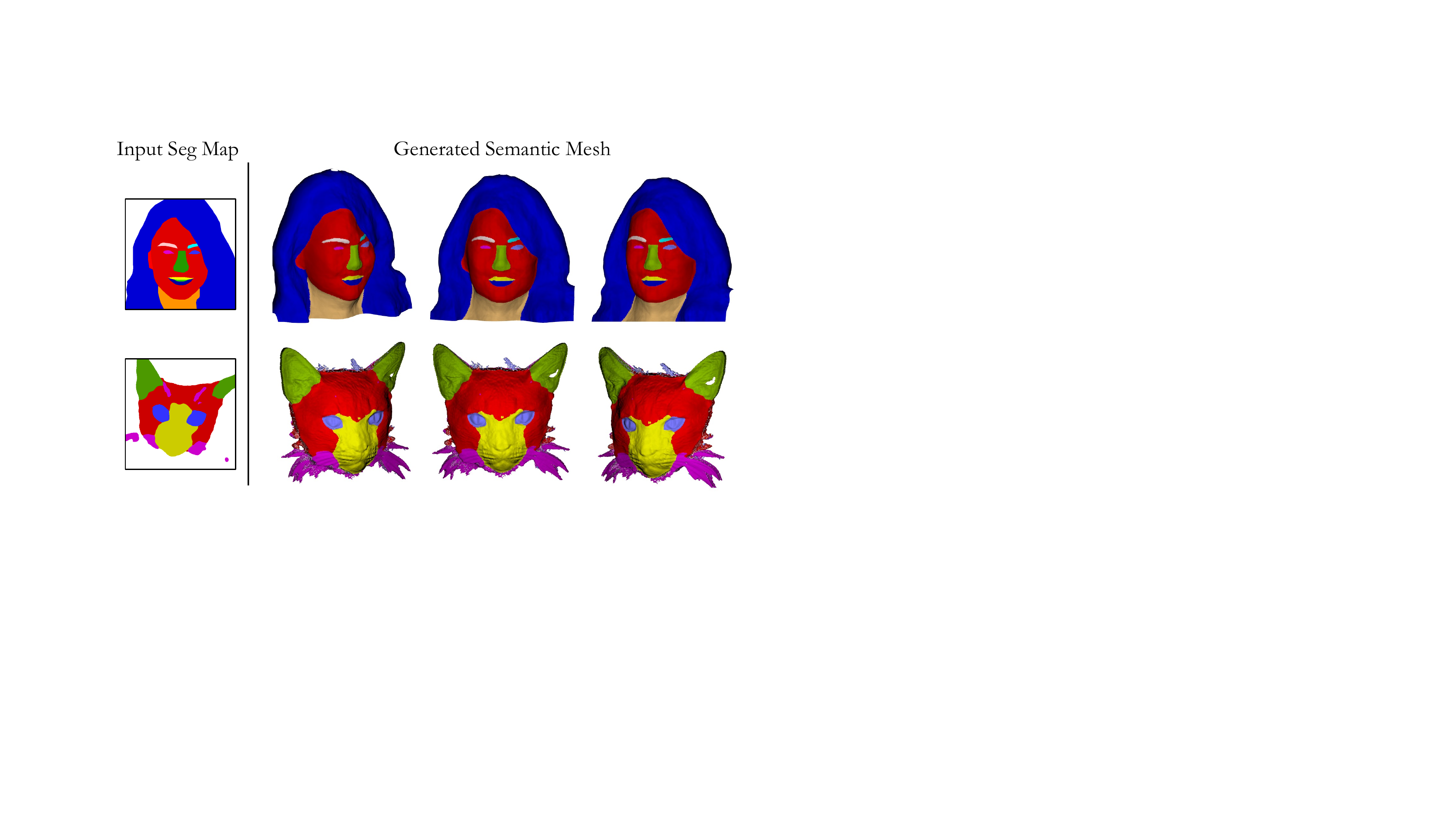}
    \caption{\textbf{Semantic Mesh.} We show semantic meshes of human and cat faces from marching cubes colored by 3D labels.}
    \vspace{-1em}
    \lblfig{mesh}
\end{figure}

%% file: figTex/fig_sample.tex
\begin{figure*}
    \centering
    \includegraphics[width=\linewidth]{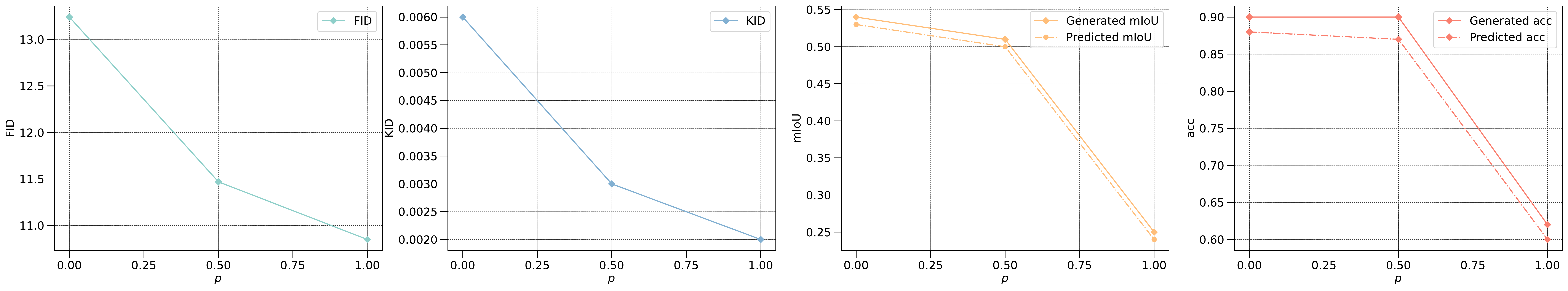}
    \vspace{-2 em}
    \caption{ We study the effect of random pose sampling probability $p$ during training. Without random poses ($p=0$), the model achieves the best alignment with input semantic maps, with reduced image quality. %
    In contrast, \textit{only} using random poses ($p=1$) achieves the best image quality, while results fail to align with input maps. We find $p=0.5$ balances the image quality and input alignment.
    }
    \vspace{-1 em}
    \lblfig{sample}
\end{figure*}

%% file: figTex/fig_edit_car.tex
\begin{figure*}[h!]
\centering
\includegraphics[width=\linewidth]{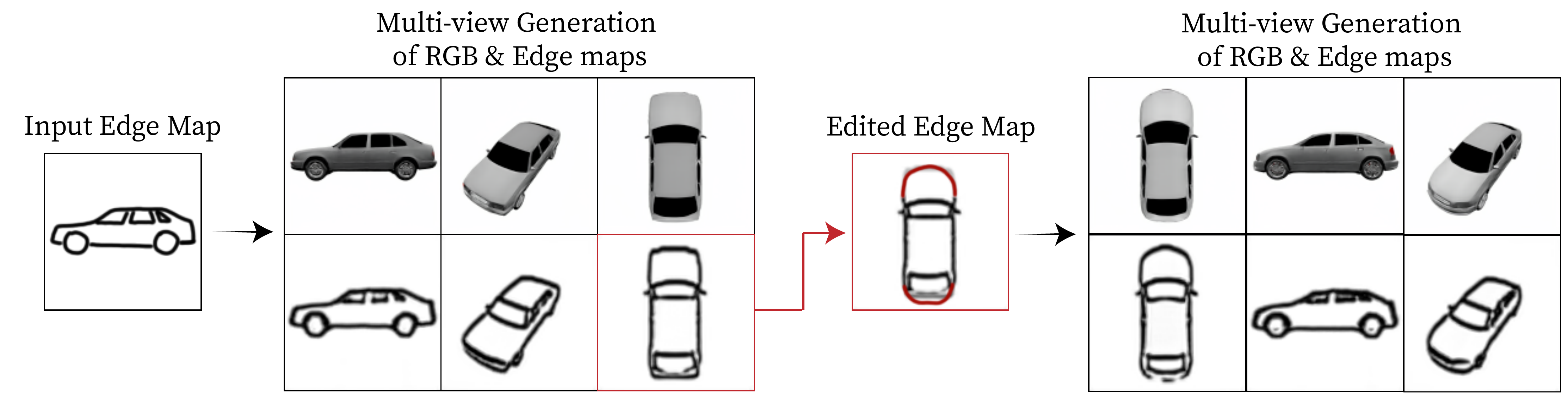}
 \vspace{-2em}
\captionof{figure}{\textbf{Cross-view Editing of Edge2Car.} 
Our 3D editing system allows users to edit label maps from any viewpoint instead of only the input view. Importantly, our feed-forward encoder allows fast inference of the latent code without GAN-inversion. Typically, a single forward pass of rendering takes only 40 ms on a single RTX A5000, which enables interactive editing. Please check our demo video on our \href{http://cs.cmu.edu/~pix2pix3D}{website}.
}

 \vspace{-1em}
\lblfig{edit_car}
\end{figure*}

%% file: figTex/fig_stylemix.tex
\begin{figure}
        \includegraphics[width=\linewidth]{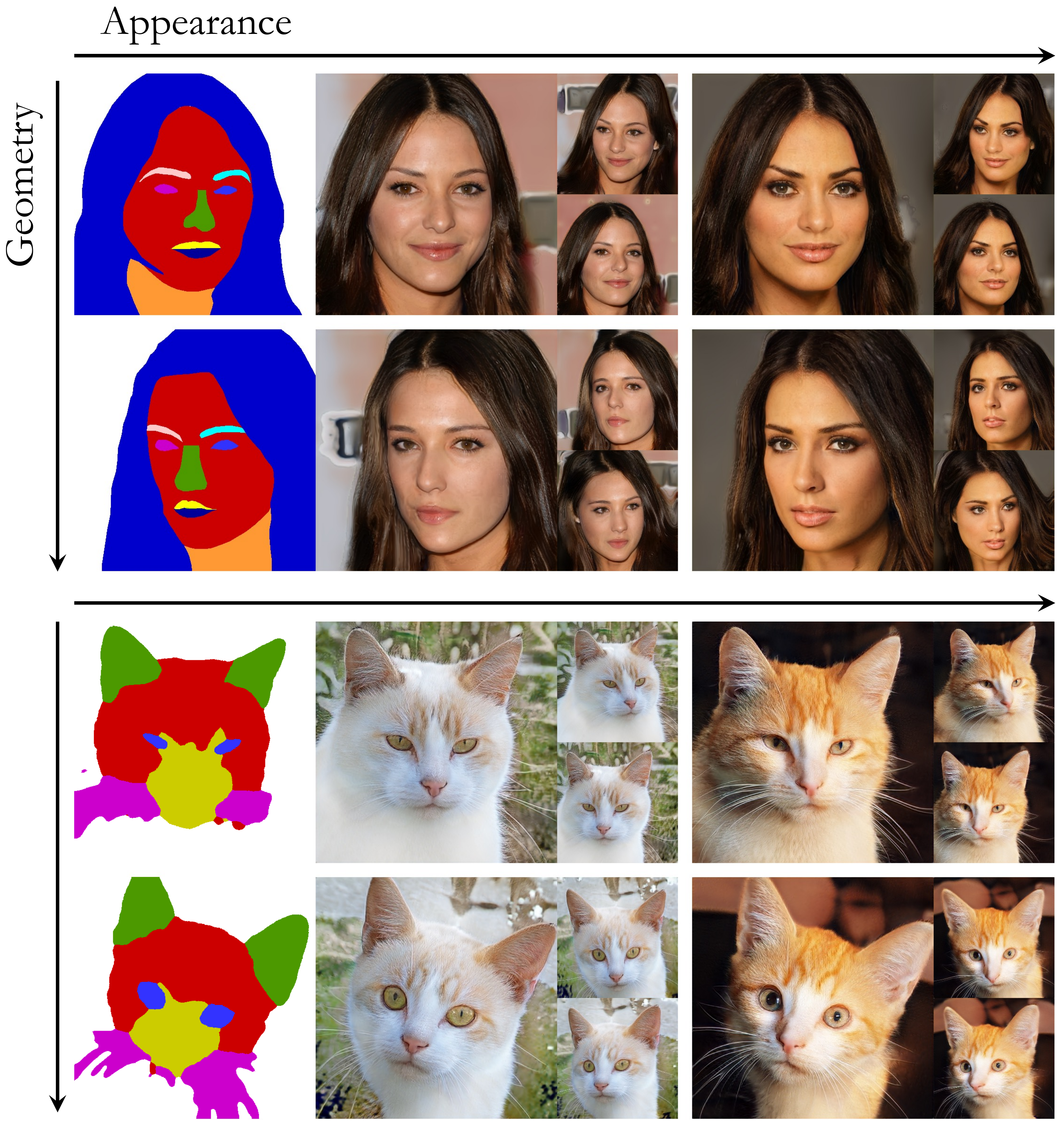}
    \vspace{-2em}
    \caption{\textbf{Multi-modal Synthesis.} The leftmost column is the input segmentation map. We use the same segmentation map for each row. We generate multi-modal results by randomly sampling an appearance style for each column.  %
    }
    \vspace{-1.5em}
    \lblfig{stylemix}
\end{figure}

%% file: figTex/fig_interp.tex
\begin{figure}[h!]
\centering
\includegraphics[width=\linewidth]{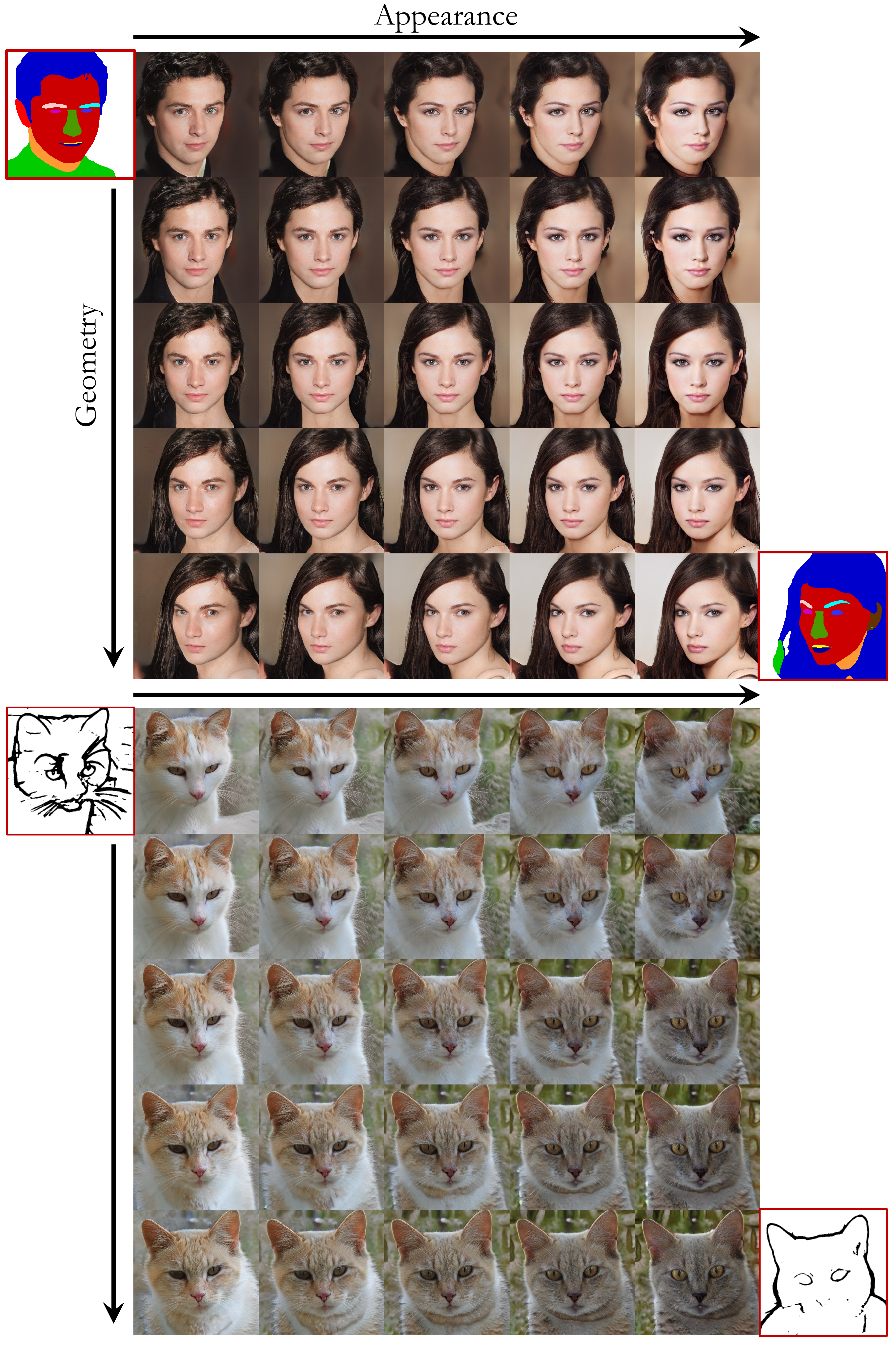}
 \vspace{-2em}
\captionof{figure}{\textbf{Interpolation.} In each $5\times5$ grid, the images at the top left and bottom right are generated from the input maps next to them.  Each row interpolates two images in label space, while each column interpolates the appearance. For camera poses, we interpolate the pitch along the row and the yaw along the column. 
 \vspace{-1.5em}
}

\lblfig{interp}
\end{figure}

%% file: sections/05_discussion.tex
\section{Discussion}
We have introduced \ourmethod, a 3D-aware conditional generative model for controllable image synthesis. Given a 2D label map, our model allows users to render images given any viewpoint. 
Our model augments the neural field with 3D labels, assigning label, color, and density to every 3D point, allowing for the simultaneous rendering of the image and a pixel-aligned label map. 
The learned 3D labels further enable interactive 3D cross-view editing. We discuss the broader impact and limitations in the appendix.

\myparagraph{Acknowledgments.} We thank Sheng-Yu Wang, Nupur Kumari, Gaurav Parmer, Ruihan Gao, Muyang Li, George Cazenavette, Andrew Song, Zhipeng Bao, Tamaki Kojima, Krishna Wadhwani, Takuya Narihira, and Tatsuo Fujiwara for their discussion and help. We are grateful for the support from Sony Corporation, Singapore DSTA, and the CMU Argo AI Center for Autonomous Vehicle Research.

%% file: sections/06_appendix.tex
\section*{Appendix}

\begin{figure*}[!ht]
    \centering
    \includegraphics[width=\linewidth]{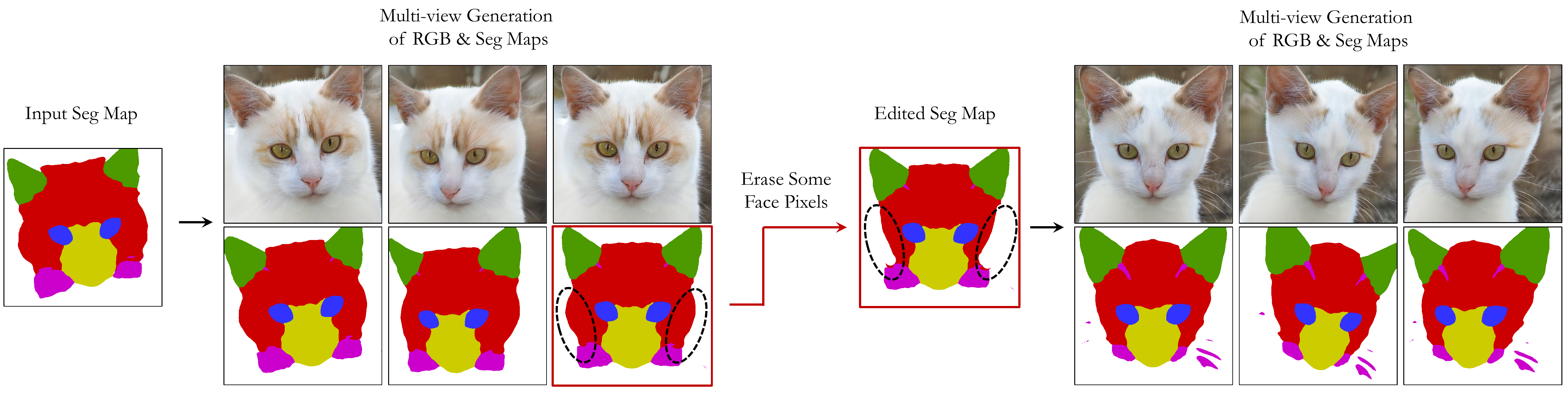}
    \captionof{figure}{\textbf{Cross-view Editing of Seg2cat.} The 3D representation can be edited from a viewpoint different than the input seg map.}
\lblfig{edit_seg2cat}
\end{figure*}

We include additional experimental results, implementation details, and the societal impact of our work. Please also view our \href{http://cs.cmu.edu/~pix2pix3D}{webpage} for our interactive editing demo video and additional visual results.

\section{Additional Experiments}
\lblsec{additional_experiments}

\myparagraph{Cross-view Editing of Seg2cat.} In addition to the edge2car editing example in the main paper, we showcase the editability of segmentation maps in \reffig{edit_seg2cat}. Note that the edited segmentation map does not have to be from the same viewpoint as the input segmentation map.

\myparagraph{Ablation study on discriminator design. } In the main paper, we introduce our Pixel-aligned Conditional Discriminator that concatenates RGB images and label maps as input. To verify the effectiveness of our discriminator design, we introduce three ablation experiments in \reftbl{abl_dis}. %
We find that our image discriminator helps improve the image quality, while our pixel-aligned conditional discriminator is crucial for the alignment.

\begin{table}[!h]
\setlength{\tabcolsep}{2pt}
\setlength{\extrarowheight}{5pt}
\renewcommand{\arraystretch}{0.8}
\small
\centering
\begin{tabular}{lcc|cc}
\toprule
\multirow{2}{*}{\textsc{CelebA-Mask}} & \multicolumn{2}{c}{\textsc{Quality}} & \multicolumn{2}{c}{\textsc{Alignment}} \\ %
               & FID $\downarrow$  & KID $\downarrow$ & mIoU $\uparrow$ & acc $\uparrow$\\ \midrule
\textsc{Ours} &  \textbf{11.54} & \textbf{0.003} & 0.51 (\textbf{0.52}) & \textbf{0.90} (0.88)  \\
\textsc{w/o Image D} & 15.32 & 0.006 & 0.51 (\textbf{0.52}) & 0.89 (0.85) \\
\textsc{w/o Conditional D} & 12.02 & 0.004 & 0.37 (0.47)  & 0.82 (0.80)\\
\textsc{w/o Pixel-Align D} & 11.94  & \textbf{0.003} & 0.41 (0.40) & 0.82 (0.81)  \\

\bottomrule
\end{tabular}
\vspace{2mm}
\caption{\textbf{Ablation Study on Discriminator Design.} %
To verify the effectiveness of our discriminator design, we introduce three ablation experiments: (1)\textsc{w/o Image D}, we remove the image discriminator and only keep the conditional discriminator that accepts the concatenation of image and segmentation maps; (2)\textsc{w/o Conditional D}, we remove the conditional discriminator and only keeps the image discriminator; (3)\textsc{w/o Pixel-Align D},  we keep both discriminators, but the conditional discriminator no longer concatenates color images as part of the input.
Our image discriminator improves the image quality, while our pixel-aligned conditional discriminator ensures alignment.
}

\lbltbl{abl_dis}
\end{table}

\myparagraph{Evaluation on Seg2Car.} We evaluate our method on an additional non-face dataset Seg2Car, where we get the segmentation model from DatasetGAN\cite{zhang21}. We show the visual and evaluation results in the \reffig{seg2car} and \reftbl{seg2car}. We find our method outperforms Pix2NeRF\cite{cai2022pix2nerf}.

\input{figTex/fig_seg2car.tex}
\input{tabTex/tab_seg2car.tex}

\reffig{sofgan_inconsistency} compares our method with SoFGAN~\cite{chen2021sofgan} regrading multi-view consistency. %
We also show our method's capability of correcting errors in the user input in \reffig{err_correction}.

\input{figTex/fig_sofgan_inconsistency.tex}

\input{figTex/fig_err_correction.tex}

\section{Implementation Details}
\lblsec{details}
\myparagraph{Class-balanced cross-entropy.} In Section 3.2 of the main text, we mentioned using class-balanced cross-entropy loss for reconstructing 2D segmentation maps. Specifically,

\begin{equation*}
    \mathcal{L}_s(\hat{\mathbf{I}}_{\mathbf{s}},\mathbf{I}_{\mathbf{s}}^+) = \mathbb{E}_n - \sum_{c=1}^C w_c \log \frac{\exp(x_{n,c})}{\sum_{i=1}^C \exp(x_{n,i})} y_{n,c}.
\end{equation*}
where $x_{n,c}$ is the semantic logits of class $c$ at location $n$ of $\mathbf{I}_{\mathbf{s}}^+$, $y_{n,c}$ is the ground-truth probability of class $c$ at location $n$ of $\hat{\mathbf{I}}_{\mathbf{s}}$, and $w_c$ is the weight of each class $c$.

In our case, 2D segmentation maps are imbalanced as skin and hair cover a lot more areas than the other classes. So we calculate $w_c$ based on the inverse frequency of the classes in the training set,
\begin{equation*}
    w_c = \sqrt{\frac{\text{\# of pixels with class $c$}}{\text{\# of all pixels}}}.
\end{equation*}

\myparagraph{Regularization.} 
As mentioned in the main text, we use non-saturating loss \cite{Goodfellow14} and R1 Regularization \cite{mescheder2018training} for GAN training following \cite{gu2022stylenerf,Karras2019stylegan2}. Specifically,

\begin{align*}
    \mathcal{L}_{\text{GAN}}(G,D) = & \mathbb{E}[f(D(G(z, \mathbf{\hat{I}_s}))] \\
    + & \mathbb{E} [f(-D(\mathbf{\hat{I}_c})) 
    + \lambda ||\nabla D(\mathbf{\hat{I}_c})||^2],
\end{align*}
where $G$ is the generator, $D$ is the discriminator, $f(u)=-\log(1+\exp(-u))$, and $\lambda = 0.5$.

\myparagraph{Hyper-parameters.} $\lambda_c = 1, \lambda_s = 5$ for edge maps, $\lambda_s = 1$ for segmentation maps, $\lambda_{D_{\bf c}}=1$, $\lambda_{D_{\bf s}}=0.1$, and $\lambda_{\text{CVC}}=1e-5$. Check our \href{http://www.cs.cmu.edu/~pix2pix3D/}{codebase} for more detailed hyperparameters.

\section{Discussion}
\myparagraph{Broader Impact.}
\lblsec{impact}
Our work allows a novice user to create 3D content more easily. The 3D outputs can be directly used in photo editing software as well as virtual reality and augmented reality applications.

Similar to recent works on data-driven 2D and 3D face synthesis, we suffer from biases in the underlying dataset. Our model is trained on CelebAMask-HQ dataset, as it provides segmentation masks that can be used as conditional input. To reduce the dataset bias, one future direction is to run our model on more diverse datasets with a pre-trained face parser. %
While our work allows for controllable 3D content generation, there may be potential misuse of the generated content. As an attempt to identify the generated content from the real photos, we run a forensics detector \cite{wang2019cnngenerated} on our generated results, and find our generated images can be detected with an accuracy of $89.77\%$, and an average precision of $99.97\%$.

\myparagraph{Usage of Existing Assets.} We use CelebAMask-HQ dataset \cite{CelebAMask-HQ}. The CelebA dataset is available for non-commercial research purposes only. All images of the CelebA dataset are obtained from the Internet. The face identities are released upon request for research purposes only. See \href{https://mmlab.ie.cuhk.edu.hk/projects/CelebA.html}{CelebA website} for details. 
We also use AFHQ Cat dataset \cite{choi2020starganv2}. This dataset is under \href{https://github.com/clovaai/stargan-v2/blob/master/LICENSE}{Creative Commons license}. Our work is also inspired by a few codebases. StyleNeRF codebase \cite{gu2022stylenerf} is under Creative Common license. StyleGAN2 codebase \cite{Karras2019stylegan2} is under the \href{https://nvlabs.github.io/stylegan2-ada-pytorch/license.html}{Nvidia Source Code License}. EG3D codebase \cite{EG3D} is under the \href{https://nvlabs.github.io/stylegan2-ada-pytorch/license.html}{Nvidia Source Code License}.

\myparagraph{Limitations.} Our current method has three major limitations. First, it mainly focuses on modeling the appearance and geometry of a single object category. Extending the method to more complex scene datasets with multiple objects is a promising next step, though defining a canonical pose for generic scenes poses a nontrivial challenge. \revision{Second, our model's generation is limited to the dataset's distribution. Our model will not follow the user input unless it is within the dataset's distribution. Incorporating diffusion models and training on more diverse datasets can potentially improve the generalization}. Finally, our model training requires camera poses associated with each training image, though our method does not require poses during inference time. Eliminating the requirement for pose information will further broaden the scope of applications. 

\section{Changelog}

\myparagraph{V2.} Add more citations in \refsec{related_work}. Fix some typos. 

\myparagraph{V1.} Initial preprint release (CVPR 2023).

%% file: figTex/fig_seg2car.tex
\begin{figure}[!h]
    \centering
    \includegraphics[width=\linewidth]{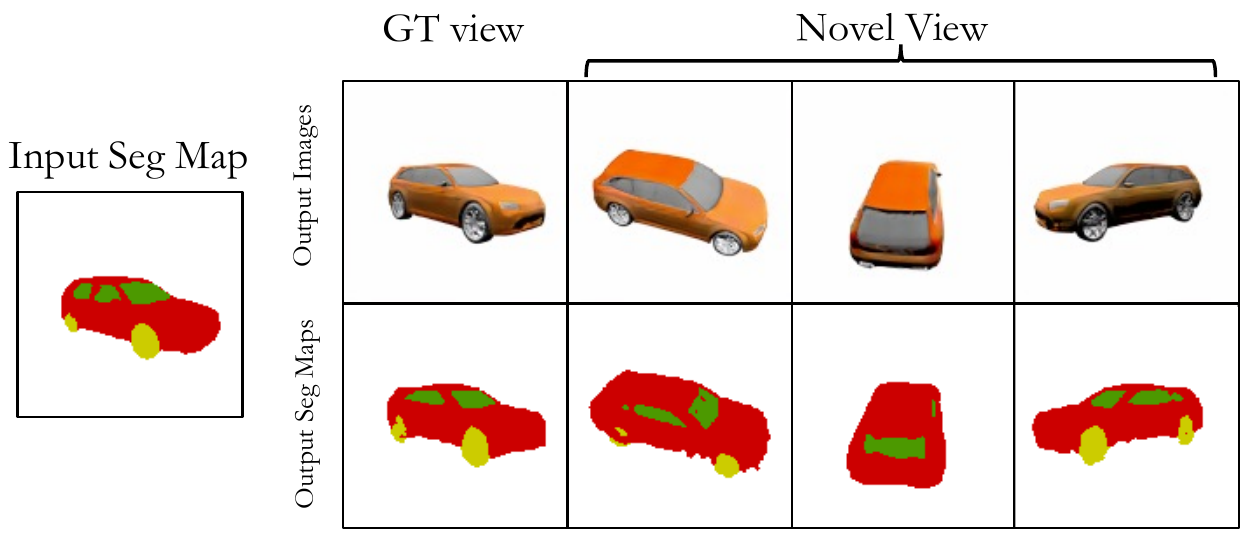}
    \caption{\textbf{Visual Results of Seg2Car.} }
    \lblfig{seg2car}
\end{figure}

%% file: tabTex/tab_seg2car.tex
\begin{table}[!h]
\setlength{\tabcolsep}{2pt}
\setlength{\extrarowheight}{5pt}
\renewcommand{\arraystretch}{0.8}
\small
\centering
\begin{tabular}{lccc|cc}
\toprule
Seg2Car & \multicolumn{3}{c}{\textsc{Quality}} & \multicolumn{2}{c}{\textsc{Alignment}} \\ %
  \textsc{Shapnet-car} & FID $\downarrow$  & KID $\downarrow$ & SG Diversity $\uparrow$ & mIoU $\uparrow$ & acc $\uparrow$  \\ \midrule
\textsc{Pix2NeRF} & 25.86 & 0.018 & 0.08 & 0.24 & 0.59 \\

\midrule
\textsc{\textbf{Ours}} & \textbf{9.35} & \textbf{0.004} & \textbf{0.14} & \textbf{0.58} & \textbf{0.88} \\

\bottomrule
\end{tabular}
\vspace{2mm}
\caption{\textbf{Seg2car Evaluation.} We compare our method with Pix2NeRF\cite{cai2022pix2nerf}. Ours performs better in all metrics.}  

\lbltbl{seg2car}
\end{table}

%% file: figTex/fig_sofgan_inconsistency.tex
\begin{figure}
    \includegraphics[width=\linewidth]{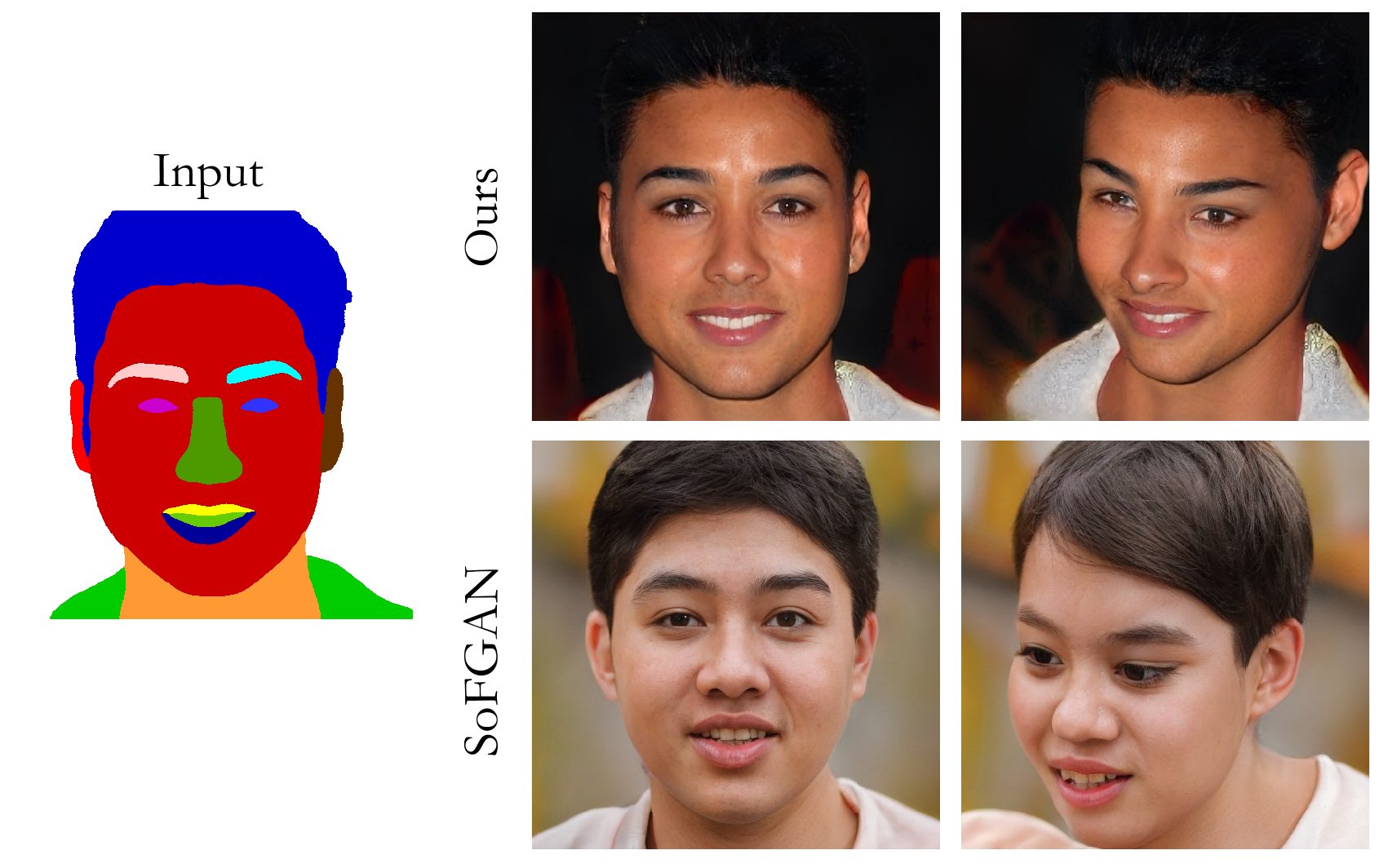}
    \vspace{-2em}
    \caption{\textbf{Multi-view Consistency}. We compare our method with SoFGAN~\cite{chen2021sofgan} regarding multi-view consistency.  Although SoFGAN can generate images from different viewpoints, their method shifts the face identity across viewpoints.  %
    In contrast, our method better preserves the identity.
    }
    \vspace{-1em}
    \lblfig{sofgan_inconsistency}
\end{figure}

%% file: figTex/fig_err_correction.tex
\begin{figure}
    \includegraphics[width=\linewidth]{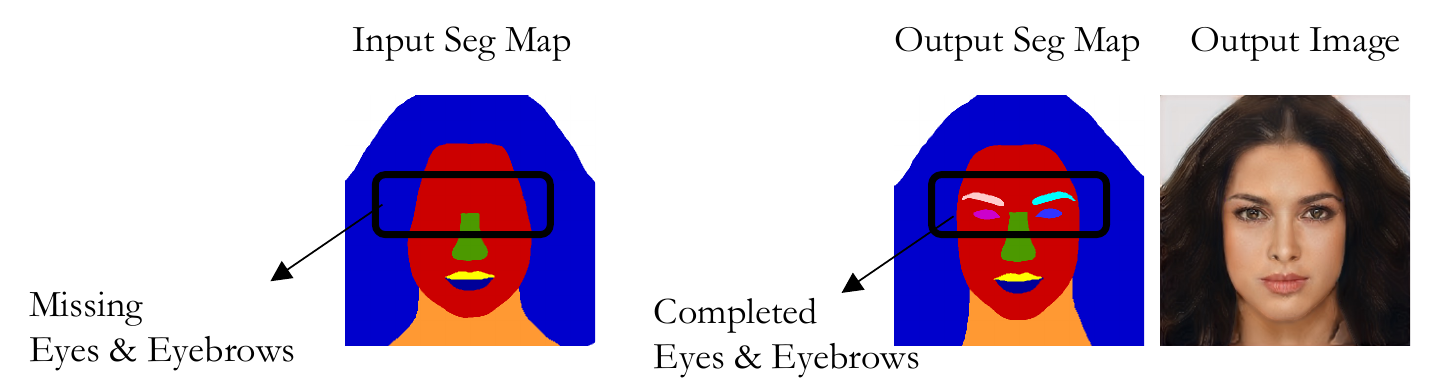}
    \vspace{-2em}
    \caption{Our model projects the user input onto the learned manifold. Even if the user input contains errors, our model is able to fix them, e.g., completing the missing eyes and eyebrows.
    }
    \vspace{-1em}
    \lblfig{err_correction}
\end{figure}